\pdfoutput=1

\documentclass[11pt]{article}

\usepackage[preprint]{acl}

\usepackage{times}
\usepackage{latexsym}

\usepackage[T1]{fontenc}

\usepackage[utf8]{inputenc}

\usepackage{microtype}

\usepackage{inconsolata}

\usepackage{graphicx}

\usepackage{amsmath, amssymb}
\usepackage{booktabs}
\usepackage{multirow}
\usepackage{array}  
\usepackage{colortbl}
\usepackage{xcolor}

\title{AsymLoRA: Harmonizing Data Conflicts and Commonalities in MLLMs Instruction Fine-tuning }

\author{
 \textbf{Xuyang Wei\textsuperscript{1,2}},
 \textbf{Chunlin Tian\textsuperscript{1}},
 \textbf{Li Li\textsuperscript{1}}
\\
 \textsuperscript{1}University of Macau
\\
 \textsuperscript{2}University of Electronic Science and Technology of China
\\
 \small{
   \textbf{Correspondence:} \href{mailto:llili@um.edu.mo}{llili@um.edu.mo}
 }
}

\begin{document}
\maketitle
\begin{abstract}
Effective instruction fine-tuning on diverse image-text datasets is crucial for developing a versatile Multimodal Large Language Model (MLLM), where dataset composition dictates the model’s adaptability across multimodal tasks. However, complex datasets often contain inherent conflicts—stemming from modality-specific optimization objectives—and latent commonalities that enable cross-task transfer, which most existing approaches handle separately. To bridge this gap, we introduce AsymLoRA, a parameter-efficient tuning framework that unifies knowledge modularization and cross-modal coordination via asymmetric LoRA: task-specific low-rank projections (matrix B) that preserve distinct adaptation pathways for conflicting objectives, and a shared projection (matrix A) that consolidates cross-modal commonalities. Extensive evaluations demonstrate that AsymLoRA consistently surpasses both vanilla LoRA, which captures only commonalities, and LoRA-MoE, which focuses solely on conflicts, achieving superior model performance and system efficiency across diverse benchmarks. \href{https://github.com/Clin0212/HydraLoRA/blob/main/MLLM-HydraLoRA/README.md}{Code.}
\end{abstract}

\section{Introduction}
\begin{figure}[!t]
    \centering
    \includegraphics[width=0.95\linewidth]{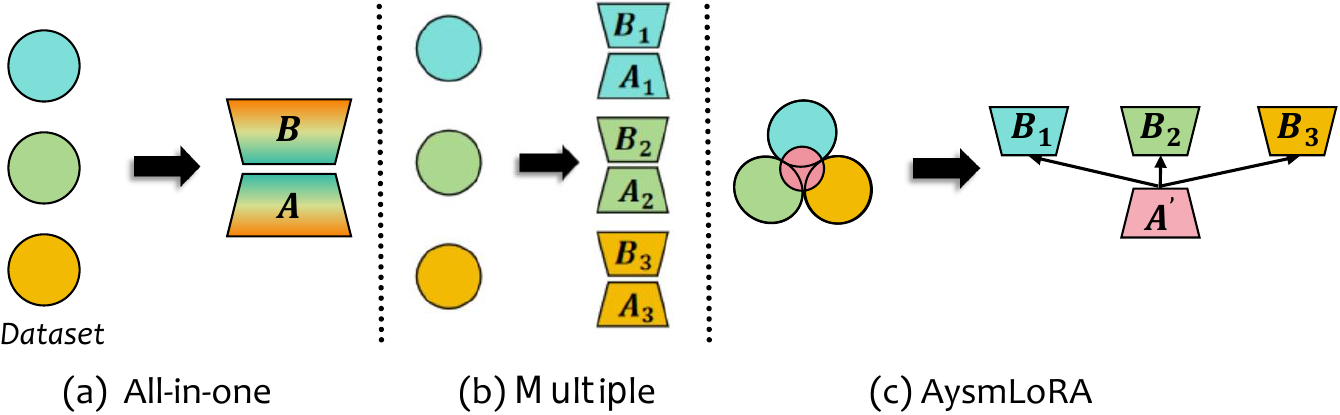}
    \caption{Illustration of LoRA architecture changes in AsymLoRA. (a) Vanilla LoRA applies a single shared adaptation across all tasks, relying solely on common synergies but introducing conflicts during fine-tuning. (b) Multiple task-specific LoRA modules mitigate interference by isolating tasks but focus only on differences, limiting generalization and increasing overhead. (c) AsymLoRA introduces an asymmetric structure with a shared A matrix for common knowledge and multiple B matrices for task-specific features, balancing generalization and efficiency.}
    \label{fig:head}
    \vspace{-1em}
\end{figure}

Multimodal Large Language Models (MLLMs) \citep{alayrac2022flamingo, huang2023language, liu2024visual, zhu2023minigpt} integrate pre-trained vision encoders with LLMs, enabling models to comprehend and generate responses based on both visual and textual inputs. To enhance their ability to handle diverse modalities and downstream tasks, MLLMs leverage instruction tuning, where models are fine-tuned on multimodal instruction-following dialogues synthesized from diverse multimodal tasks. Parameter-Efficient Fine-Tuning (PEFT) techniques \citep{houlsby2019parameter,liu2021p}, such as Low-Rank Adaptation (LoRA) \citep{hu2022lora}, enhance adaptability by injecting small trainable components into the model, significantly reducing trainable parameters while preserving or even improving task performance. However, directly applying LoRA in multi-task learning (see Figure \ref{fig:head} (a)) can lead to conflicting optimization objectives, where task-specific adaptations interfere with each other, potentially canceling out useful task-specific updates. 
To address this, Mixture-of-Experts (MoE) \citep{jacobs1991adaptive} extends LoRA by introducing specialized modules (see Figure \ref{fig:head} (b)) that learn task-specific knowledge \citep{lin2024moe,chen2024llava}, improving alignment across diverse modalities. However, multi-task datasets inherently contain both conflicts—arising from modality-specific optimization goals—and latent commonalities that facilitate cross-task transfer \citep{tian2024hydralora}, which cannot be effectively tackled in isolation. 

In this work, We propose AsymLoRA, an asymmetric LoRA architecture designed for MLLM instruction fine-tuning. Unlike vanilla LoRA, which applies a single pair of low-rank decomposed matrices to the Transformer MLP layers, AsymLoRA introduces a shared A matrix for capturing common knowledge and task-specific B matrices for independent task adaptation (see Figure \ref{fig:head} (c)). During inference, AsymLoRA dynamically selects task-specific B matrices in a MoE manner, effectively balancing commonalities and conflicts across multi-task datasets to enhance both efficiency and performance. Through extensive experiments on diverse data configurations, we demonstrate that AsymLoRA effectively mitigates conflicts between instruction datasets while leveraging their shared knowledge, achieving superior performance and efficiency with fewer parameters. We summarize our primary contributions as follows:
\begin{itemize}
    \item Based on an advanced MLLM model and large-scale
datasets, identify the inherent conflicts and commonalities in instruction fine-tuning MLLMs on mixtures datasets.
\item We propose AsymLoRA, an asymmetric LoRA architecture that addresses conflicts through task-specific B matrices while capturing commonalities via a shared A matrix.
\item Extensive experiments across multiple benchmarks validate that AsymLoRA consistently outperforms both vanilla LoRA and LoRA-MoE in terms of both performance and efficiency across various data configurations.
\end{itemize}

\section{Background and Motivation}
\paragraph{Low-Rank Adaptation}\citep{hu2022lora} is an efficient fine-tuning technique for large pre-trained models, introducing small low-rank matrices (A and B) that can be applied to arbitrary linear layers. Formally, for a linear transformation \( h = Wx \) with input \( x \in \mathbb{R}^{d_{i}} \) and weight \( W \in \mathbb{R}^{d_{o} \times d_{i}} \), LoRA learns a low-rank decomposed update:  

\begin{equation}
    y' = y + \Delta y = Wx + BAx
\end{equation}

where \( y \in \mathbb{R}^{d_{o}} \) is the output, and \( A \in \mathbb{R}^{r \times d_{i}} \), \( B \in \mathbb{R}^{d_{o} \times r} \) are low-rank matrices with \( r \ll \min(d_{o}, d_{i}) \) as the chosen rank. Typically, B is initialized to zeros, while A follows Kaiming Uniform initialization \citep{he2015delving}. During fine-tuning, only A and B are updated, keeping the original model parameters frozen, thus significantly reducing computational overhead.

\paragraph{Observation I: Knowledge contains inherent conflicts.} 
Instruction fine-tuning on diverse image-text datasets is crucial for enhancing the performance of MLLMs, where the configuration of training data plays a pivotal role. However, we observe that when instruction data from different domains are combined, inherent conflicts between domain-specific optimization objectives become inevitable. These conflicts often lead to a significant performance drop in certain domains compared to fine-tuning on a single-domain dataset.
As shown in Table \ref{tab:obs}, we fine-tune LLaVA \citep{liu2024visual} using instruction data from two distinct domains—Visual Question Answering (VQA) \citep{antol2015vqa} and Generative (Gen, including LLaVA-15k \citep{liu2024visual}, VQG \citep{mostafazadeh2016generating})—both separately and in combination. While in certain cases, such as the “Yes/No” (78.08\%) and “Other” (36.41\%) tasks in VizWiz, the increased data volume from mixing domains leads to performance gains, benchmark evaluations generally reveal that naively combining instruction data from different domains significantly degrades performance. For instance, on the TextQA \cite{singh2019towards} test set, the Gen-only model achieves 54.25\%, whereas the VQA+Gen mixed model drops to 43.25\%. These results underscore that the scalability of multi-domain instruction tuning is inherently constrained by dataset conflicts, highlighting the need for more sophisticated strategies to ensure effective multi-domain adaptation.

\begin{table}[!t]
    \centering
    \resizebox{\linewidth}{!}{
    \begin{tabular}{c|c|cccc|cc}
    \bottomrule[1.5pt]
         & &\multicolumn{4}{c|}{\textbf{VizWiz(\%)}} & \multicolumn{2}{c}{\textbf{MME}}  \\
         \multirow{-2}{*}{\textbf{Schemes}} & \multirow{-2}{*}{\textbf{TextVQA (\%)}}  & \textbf{Other}& \textbf{Unanswerable}& \textbf{Yes/No}&\textbf{Number} &\textbf{Perception}& \textbf{Cognition} \\
         \toprule[0.75pt]
         \multicolumn{8}{c}{Single LoRA} \\
         \toprule[0.75pt]
    VQA &\cellcolor{green!5}{38.08}& \cellcolor{green!30}{31.81}& \cellcolor{green!80}{39.19} & \cellcolor{green!5}{73.64}& \cellcolor{green!30}{22.76} & \cellcolor{green!5}{1152.46} & \cellcolor{green!5}{224.64}\\
    
    Gen &\cellcolor{green!80}{54.80}& \cellcolor{green!5}{27.13}& \cellcolor{green!80}{76.32}& \cellcolor{green!30}{73.9} & \cellcolor{green!80}{30.24} & \cellcolor{green!80}{1255.63} & \cellcolor{green!80}{296.07}\\
    
    VQA+Gen & \cellcolor{green!30}{43.25} & \cellcolor{green!80}{36.41} & \cellcolor{green!30}{28.91} & \cellcolor{green!80}{78.08}& \cellcolor{green!5}{22.28} & \cellcolor{green!30}{1203.66} & \cellcolor{green!30}{268.92}\\
    
    \toprule[0.75pt]
    \multicolumn{8}{c}{Multiple LoRA (MoE)} \\
    \toprule[0.75pt]
     VQA+Gen &\cellcolor{green!80}{54.05}& \cellcolor{green!80}{39.66}& \cellcolor{green!5}{24.8}& \cellcolor{green!80}{82.15}& \cellcolor{green!80}{33.98} & \cellcolor{green!80}{1454.37} & \cellcolor{green!80}{324.64}\\
    \bottomrule[1.5pt] 
    \end{tabular}}
    \caption{Performance of fine-tuning MLLM (LlaVA-1.5-7B \citep{liu2024visual}) on benchmarks (TextVQA\citep{singh2019towards}, VizWiz\citep{bigham2010vizwiz}, MME\citep{fu2024mmecomprehensiveevaluationbenchmark}) across different instruction data domains.}
    \label{tab:obs}
    \vspace{-1em}
\end{table}

\paragraph{Observation II: Knowledge contains latent commonalities.}
While modularity is essential, knowledge across different tasks is often complementary, allowing shared learning to enhance performance. MLLMs should be capable of capturing, integrating, and evolving with diverse knowledge from multiple sources and perspectives, enabling collaborative learning across various domains. As shown in Table \ref{tab:obs}, we fine-tune separate LoRA modules on different instruction datasets and treat each LoRA module as an expert, dynamically combining them via Mixture-of-Experts (MoE) based on randomized inputs. MoE-LoRA achieves the highest accuracy in certain cases, such as MME scores for Perception (1454.37) and Cognition (324.64), as well as the “Other” (39.66\%) and “Yes/No” (82.15\%) categories in the VizWiz benchmark. These results demonstrate that LoRA-MoE effectively leverages modularity by using task-specific modules, reducing interference, and enhancing performance. However, in other scenarios, MoE-LoRA underperforms compared to single-domain fine-tuning. For example, on TextVQA, it falls short of the Gen-only fine-tuned model, and in the Unanswerable task of VizWiz, it even records the worst performance. This suggests that focusing solely on task differences while overlooking latent commonalities can limit the full potential of the data, ultimately reducing model effectiveness. A balanced approach that respects both task-specific adaptations and shared knowledge is necessary for optimal performance.

\section{Methodology}

\subsection{AsymLoRA Architecture}
\begin{figure}[!t]
    \centering
    \includegraphics[width=1.0\linewidth]{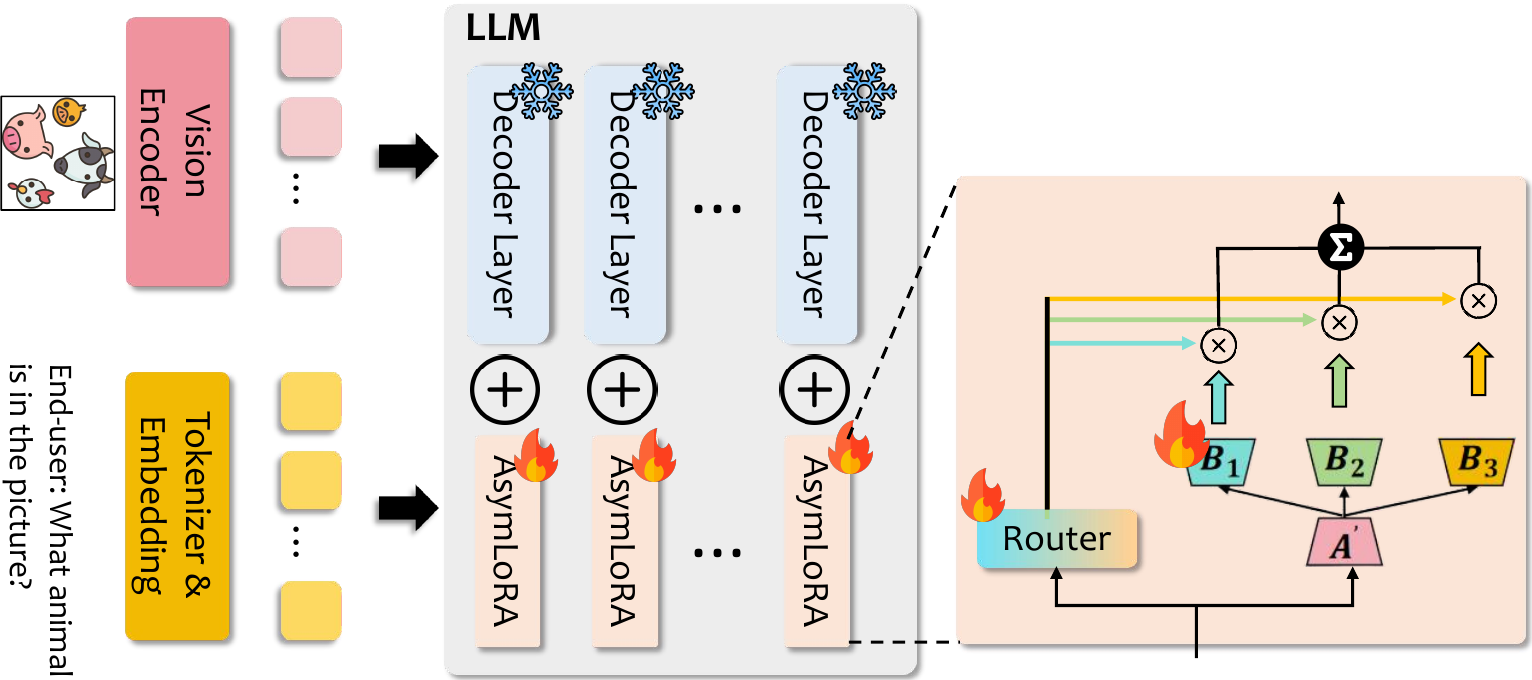}
    \caption{Architecture and workflow of AsymLoRA. The shared low-rank matrix $A$ captures global knowledge across tasks, while task-specific low-rank matrices $B_i$ enable independent adaptation for each task}
    \label{fig:framework}
    \vspace{-1em}
\end{figure}

As illustrated in Figure \ref{fig:framework}, AsymLoRA introduces an asymmetric design for efficient MLLM instruction fine-tuning, addressing the limitations of traditional symmetric LoRA approaches. Unlike conventional methods that apply both $A$ and $B$ matrices uniformly across all tasks, AsymLoRA maintains a shared low-rank matrix $A$ to capture common knowledge while introducing task-specific low-rank matrices $B_i$ to enable specialized adaptation.
Formally, given a dataset $D = \{D_1, D_2, \dots, D_N\}$ where each $D_i$ corresponds to a subtask $T_i$, our objective is to optimize shared parameters $A$ and task-specific parameters $B_i$ to minimize the task-specific loss $L_i$ for each $T_i$:  
\begin{equation}
    \min_{A, B_i} \sum_{i=1}^{N} L_i(T_i; A, B_i).
\end{equation}
Shared matrix $A$ facilitates knowledge transfer across tasks, reducing the number of trainable parameters and enhancing generalization, while the $B_i$ matrices provide targeted adaptations, mitigating interference between conflicting tasks.

\subsection{Mixture of AsymLoRA Experts}
To further enhance adaptability and performance in multi-task learning, we extend AsymLoRA with a Mixture of Experts (MoE) mechanism. In this approach, multiple experts share a common low-rank matrix $A$, representing global knowledge across tasks, while each expert is associated with a distinct set of task-specific matrices $\{B_1^j, B_2^j, \dots, B_N^j\}$, where $j$ denotes the expert index. Given a task $T_i$, a gating network dynamically selects the most suitable expert based on task-specific input features. The gating network assigns a weight $w_j$ to each expert $j$ and computes the final task-specific transformation as:
\begin{equation}
    \text{Output}_i = f(x; A, \sum_{j} w_j B_i^j ).
\end{equation}
This design ensures that while the shared matrix $A$ provides consistent general knowledge, the expert-specific $B_i^j$ matrices enable flexible task-specific adaptation. By dynamically selecting the most appropriate expert, AsymLoRA with MoE improves the model’s ability to handle diverse tasks while minimizing interference, leading to more robust and efficient multi-task learning.

\section{Experiments}
\subsection{Experiment Setting}

\paragraph{Model.} Following LLaVA-1.5 \citep{liu2024visual} utilizing CLIP ViT-L \citep{radford2021learning} as the vision encoder with a 336×336 input resolution and a 14×14 patch size. A two-layer MLP adapter processes the 576 tokens extracted from ViT. The language model is Vicuna-7B \citep{vicuna2023}, with both ViT and Vicuna weights kept frozen throughout training. Unless stated otherwise, LoRA is applied to the LLM with a rank of 32, the number of B matrix is initialized N = 3.

\begin{table*}[!t]
    \centering
        \centering
        \resizebox{0.85\linewidth}{!}{
            \begin{tabular}{c|c|c|cc|cccc}
                \bottomrule[1.5pt]
                &  & &\multicolumn{2}{c|}{\textbf{MME}} & \multicolumn{4}{c}{\textbf{GQA}} \\
                \multirow{-2}{*}{\textbf{Schemes}}  & \multirow{-2}{*}{\textbf{TextVQA} (\%)} & \multirow{-2}{*}{\textbf{MM-Vet(\%)}} & \textbf{Perception} & \textbf{Cognition} & \textbf{Binary} (\%) & \textbf{Open} (\%) & \textbf{Acc.} (\%) & \textbf{Dis.} ($\downarrow$) \\ \toprule[0.75pt]
                LoRA     & 36.43 & \underline{31.5} & 911.3 & \underline{278.21} & 12.04 & 8.32 & 10.03 & 3.62 \\ 
                MoE-LoRA & \underline{53.33} &31.2 & \underline{1121.88} & 270.01 &  \underline{66.90} &  \underline{42.73} & \underline{53.82} &  \underline{1.58}\\ 
                 \textit{AsymLoRA} & \textbf{55.51} & \textbf{31.8} & \textbf{1327.93} &\textbf{287.14} & \textbf{75.23} & \textbf{46.35} & \textbf{59.60} & \textbf{1.50}\\ 
                \toprule[0.75pt]
            \end{tabular}}
    \caption{Evaluation results for fine-tuning LlaVA on single domain (Conversation-58k).}
    \label{tab:Conversation}
    \vspace{-1em}
\end{table*}

\paragraph{Dataset and Benchmarks.}
We evaluate our model in single-domain and multi-domain settings across diverse multimodal tasks.
1) For the single-domain setting, training is conducted on \textbf{Conversation\_58k}, a dataset with 58,000 conversational examples for dialogue-based learning, and \textbf{LLaVA\_v1\_5\_mix665k}, a large-scale mixed dataset for multimodal training.
2) In the multi-domain setting, training combines\textbf{ VQA, LLaVA-15k, and VQG}, where VQA is a large-scale dataset for open-ended visual question answering, LLaVA-15k consists of 15,000 vision-language task samples, and VQG facilitates natural question generation for conversational AI. 
Evaluation is performed on multiple benchmarks, including \textbf{MME} \citep{fu2024mmecomprehensiveevaluationbenchmark} (multimodal integration and reasoning), \textbf{GQA} \citep{hudson2019gqa} (scene graph-based VQA), \textbf{MM-Vet} \citep{yu2024mmvetevaluatinglargemultimodal} (Integrated Capabilities of MLLM) \textbf{VizWiz} \citep{bigham2010vizwiz} (real-world VQA with noisy images), and \textbf{TextVQA} \citep{singh2019towards} (requiring integration of textual and visual information).

\subsection{Overall Performance.}
We present the experimental results of AsymLoRA and competing baselines across three evaluation settings: single-domain conversation tasks (Table \ref{tab:Conversation}), single-domain general tasks (Table \ref{tab:mix60k}), and multi-task domain settings (Table \ref{tab:multi-domain}). The results demonstrate that AsymLoRA consistently outperforms all other schemes, validating its effectiveness in multimodal instruction fine-tuning.

\textbf{Performance in Single-Domain Tasks.}
In Table \ref{tab:Conversation}, AsymLoRA achieves a TextVQA score of 55.51\%, surpassing MoE-LoRA (53.33\%) and significantly outperforming LoRA (36.43\%), indicating its superior ability to integrate textual and visual cues. On the MME benchmark, AsymLoRA achieves the highest Perception (1327.93) and Cognition (287.14) scores, outperforming MoE-LoRA (1121.88, 270.01) and LoRA (911.3, 278.21), showcasing its enhanced multimodal reasoning and feature extraction capabilities. Furthermore, in GQA, AsymLoRA attains the highest accuracy (59.60\%) while minimizing distribution shift (1.50), highlighting its robust generalization in structured reasoning tasks.
Similarly, in Table \ref{tab:mix60k}, AsymLoRA achieves consistent gains in the VizWiz benchmark, leading in Unanswerable (81.70\%) and Number (43.33\%), with the highest overall average score (51.31\%), demonstrating its robustness in real-world visual question answering. This improvement is attributed to AsymLoRA’s asymmetric design, which effectively balances common knowledge (A matrix) and task-specific adaptation (B matrices), mitigating conflicts between different domains while retaining the strengths of both LoRA and MoE-LoRA.

\textbf{Robust Multi-Task Adaptation.}
In the multi-task domain setting (Table \ref{tab:multi-domain}), AsymLoRA continues to outperform competing methods across Perception, Cognition, and reasoning-based benchmarks, demonstrating its superior capability in handling diverse multimodal challenges. Specifically, it achieves the highest TextVQA score (54.25\%) and VizWiz average (38.10\%), improving upon MoE-LoRA (53.84\%, 37.44\%) and LoRA (43.25\%, 36.00\%). These results highlight AsymLoRA’s advantage in adapting dynamically to different domains while preserving effective knowledge transfer across tasks, leading to overall superior multi-task performance.


\begin{table}[!t]
    \centering
    \resizebox{\linewidth}{!}{
    \begin{tabular}{c|cccc|c|cc}
    \bottomrule[1.5pt]
        & \multicolumn{5}{c|}{\textbf{VizWiz (\%)}} &\multicolumn{2}{c}{\textbf{MME}} \\
        \multirow{-2}{*}{\textbf{Schemes}}  & \textbf{Unanswerable} & \textbf{Yes/No} & \textbf{Number}  & \textbf{Other} & \textbf{Average} & \textbf{Perception} & \textbf{Cognition}  \\ \toprule[0.75pt]
        LoRA     & 55.32 & 75.27 & \underline{39.43} & \textbf{39.22}& 45.12 & \underline{1446.44} & 262.5  \\
        MoE-LoRA & \underline{72.47} & \textbf{75.79} & 39.27 & \underline{38.75} & \underline{49.43} & 1446.31 & \underline{306.07} \\
        \textit{AsymLoRA} & \textbf{81.70} & \underline{75.53} & \textbf{43.33} & 37.79 & \textbf{51.31} & \textbf{1489.19} & \textbf{367.14}  \\
      \toprule[0.75pt]
    \end{tabular}}
    \caption{Evaluation results for fine-tuning LlaVA on single domain (LLaVA-mix665k).}
    \label{tab:mix60k}
\end{table}

\begin{table}[!t]
    \centering
    \resizebox{\linewidth}{!}{
    \begin{tabular}{c|cccc|c|c}
    \bottomrule[1.5pt]
    & \multicolumn{5}{c|}{\textbf{VizWiz (\%)}} &\\
    \multirow{-2}{*}{\textbf{Schemes}}  & \textbf{Unanswerable} & \textbf{Yes/No} & \textbf{Number}  & \textbf{Other} & \textbf{Average} & \multirow{-2}{*}{\textbf{TextVQA (\%)}}   \\ \toprule[0.75pt]
    LoRA & \textbf{28.91} & 78.08 & 22.28 & 36.41 & 36.00 & 43.25 \\
    MoE-LoRA & \underline{24.80} & \textbf{82.15} & \underline{33.98} & \underline{39.66} & \underline{37.44} & \underline{53.84} \\
    \textit{AsymLoRA} & 23.65 & \underline{78.10} & \textbf{34.80} & \textbf{41.35} & \textbf{38.10} & \textbf{54.25} \\
    \toprule[0.75pt]
    \end{tabular}}
        \caption{Evaluation results for fine-tuning LlaVA on multi-task domain (VQA, LLaVA-15k, VQG).}
    \label{tab:multi-domain}
\end{table}

\section{Conclusion}
This work presents AsymLoRA, a parameter-efficient tuning framework that balances modality-specific conflicts and cross-task commonalities in MLLM fine-tuning. By leveraging task-specific B matrices for adaptation and a shared A matrix for knowledge transfer, AsymLoRA outperforms vanilla LoRA and LoRA-MoE, achieving superior performance and efficiency across benchmarks. These results highlight its effectiveness as a scalable solution for multi-task instruction fine-tuning.

\section{Limitations}
Despite its advancements, AsymLoRA has several limitations. First, while the framework demonstrates strong performance on established multimodal benchmarks, its generalizability to entirely unseen modalities or highly specialized domains (e.g., medical imaging or low-resource languages) remains unverified. Second, the dynamic selection of task-specific B matrices via MoE introduces non-trivial computational overhead during inference, which may limit deployment in latency-sensitive scenarios.  Finally, the theoretical underpinnings of how the shared A matrix consolidates cross-modal commonalities warrant deeper exploration to better guide future architectural designs. Addressing these limitations could further enhance the framework’s practicality and robustness.
\bibliography{custom}




\end{document}